%% file: acl2019_main.tex
\documentclass[11pt,a4paper]{article}
\usepackage[hyperref]{acl2019}
\usepackage{times}
\usepackage{latexsym}

\usepackage{graphicx}
\usepackage{url}
\usepackage{xcolor}
\usepackage[normalem]{ulem}
\usepackage{multirow}
\usepackage{booktabs}
\usepackage{dirtytalk}
\usepackage{enumitem}

\aclfinalcopy 

\title{From Receptive to Productive: Learning to Use Confusing Words \\through Automatically Selected Example Sentences}

\author{Chieh-Yang Huang$^1$, Yi-Ting Huang$^2$, Mei-Hua Chen$^3$ and Lun-Wei Ku$^2$ \\
  $^1$ IST, Pennsylvania State University, USA, $^2$ IIS, Academia Sinica, Taiwan \\
  $^3$ FLLD, Tunghai University, Taiwan \\
  $^1$ \texttt{chiehyang@psu.edu}, $^2$ \texttt{\{ythuang, lwku\}@iis.sinica.edu.tw} \\
  $^3$ \texttt{mhchen@thu.edu.tw} \\
}

\date{}

\begin{document}
\maketitle
\begin{abstract}
Knowing how to use words appropriately has been a key to improving language
proficiency. Previous studies typically discuss how students learn receptively
to select the correct candidate from a set of confusing words in the
fill-in-the-blank task where specific context is given. In this paper, we go
one step further, assisting students to learn to use confusing words
appropriately in a productive task: sentence translation.
We leverage the GiveMeExample system, which suggests example sentences for each confusing word, to achieve this goal. In this study,
students learn to differentiate the confusing words by reading the example
sentences, and then choose the appropriate word(s) to complete
the sentence translation task. Results show students made substantial
progress in terms of sentence structure. In addition, highly proficient students better managed to learn
confusing words. In view of the influence of the first language on learners, we further propose an effective approach to improve the quality of the
suggested sentences.

\end{abstract}

\section{Introduction}

In second or foreign language learning, learning synonyms is not uncommon in
vocabulary learning~\cite{hashemi2005attribute,webb2007effects}. However, clear
differentiation and proper use of near-synonyms poses a challenge to many
language learners~\cite{laufer1990ease,tinkham1993effect,waring1997negative}.
Researchers have investigated language learners' lexical use problems,
e.g.,~\cite{chen2011factors,hemchua2006analysis,yanjuan2014bnc,laufer1990ease,tinkham1993effect,waring1997negative,yeh2007online,zughoul1991lexical}
and suggested that discriminating among semantically similar items presents
difficulties for learners~\cite{laufer1990ease}. For example, 
Zughoul~\shortcite{zughoul1991lexical} analyzed the writings of Arab EFL college
students and found that misapplication of near-synonyms was the most common
type of word choice error made by his students. Likewise, Hemchua and 
Schmitt~\shortcite{hemchua2006analysis} investigated lexical error types in the writings 
of Thai college students and found that the use of near-synonyms was the most
common error made by their students. 

Learners are prone to assuming that synonyms behave identically in all 
contexts~\cite{martin1984advanced}. Actually, even though two words may share similar
meanings, they may not be fully substitutable in certain 
scenarios~\cite{edmonds2002near,karlsson2014advanced,liu2014l2,martin1984advanced,webb2007effects}.
Synonyms are highly likely to confuse learners~\cite{martin1984advanced}. For
example, both \textit{emphasis} and \textit{stress} describe \say{special
attention or importance}. The verbs \textit{lay}, \textit{place}, and
\textit{put} can collocate with \say{emphasis on} and \say{stress on}; however,
\say{place stress on} is a rare expression (it occurs only once in the British
National Corpus). For ESL/EFL learners, correct word usage necessitates not only 
knowledge of the meaning of a word but also knowledge of its paradigmatic 
and syntagmatic association. Without usage
information, synonyms \say{usually leave the student 
mystified}~\cite{martin1984advanced}. Verbs \textit{construct} and \textit{establish}
illustrate the fact that synonyms do not always have the same 
collocates~\cite{webb2007effects}. Although both words share the same meaning of
\say{build}, in practice, they are not interchangeable in the
collocations \say{establish contact} and \say{construct system}. 
Learners must grasp the collocational and syntactic differences to use synonyms
effectively in a productive mode~\cite{martin1984advanced}.

For language learners, to facilitate the use of near-synonyms, confusing words, or collocations, 
it is not enough to just learn the senses of a single confusing word. This has led 
to the design of learning materials such as thesauri and dictionaries for confusing 
and easily-misused words~\cite{room1988dictionary,ragno2016use}. Although the
information these reference tools provide is appropriate and instructive,
the contents~-- especially example sentences~-- are neither rich nor
constantly updated. 

In view of this, artificial intelligence techniques recently have been widely
applied to assist language learning. Applications such as grammar 
correction~\cite{ng2014conll,napoles2017systematically} and essay 
scoring~\cite{alikaniotis2016automatic,dong2016automatic,zhang2018co} are relatively
mature. 
Research on the lexical substitution \cite{mccarthy-navigli-2007-semeval, mccarthy2009english, mihalcea2010semeval, melamud2015simple} and the detection and correction of collocation errors \cite{futagi2010effects, alonso2014towards} have also shown the potential of helping ESL learn similar words, near-synonyms or synonyms.
Lexical substitution task try to determine a substitute for a word in a context and preserving its meaning and is possible to help language learners understand the correct meaning of a target word by selecting a lexical substitute.  
The detection and correction, on the other hand, is an inevitable assistance for ESL learners since, as we know, collocation error is one of the most common lexical misuse problem.
However, as interpretation is still challenging for AI models,
especially deep learning models~\cite{ribeiro2016should,doshi2017towards},
there are fewer applications for tasks involving comparisons and 
explanations, which is the key to learning confusing words.

GiveMeExample~\cite{huang2017towards} is one of the few systems. It offers students
suggestions of example sentences for confusing words and helps them to choose proper words 
for fill-in-the-blank multiple-choice questions.
GiveMeExample aims to provide opportunities for learners to self-learn the
nuances between confusing words by comparing and contrasting the suggested
example sentences. However, the fill-in-the-blank multiple-choice format  
has its limitations. First, it decreases learning efficiency:
students look for hints (such as prepositions or collocations)
from the example sentences to match the words adjacent to the blank instead of
reading and comparing these example sentences thoroughly.  Also,
as answering multiple-choice questions is a discriminative task, students attempt
to select the most possible candidate among all choices instead of
learning to properly use the confusing words in question.

To improve the learning effect, we adopt GiveMeExample but deploy it using a carefully designed sentence translation task. 
Studies \cite{uzawa1996second, prince1996second, laufer2008form} have investigated the effect of using translation tasks in language learning. 
With the integration of the translation task, learners were asked to produce a second language (L2) text conditioned on a given first language (L1) sentence. 
It is one of effective ways to learn word usage by producing a good translation.
In other words, we intentionally move from a receptive to a productive learning task. 
Generating sentences using confusing words requires a better understanding of the words: with this task we hope to discover how to better assist language learners to learn to differentiate confusing words.  

\input{tables/clarification_example.tex}

\section{Automatic Example Sentence Selection}
In this study, we seek to use the GiveMeExample system \cite{huang2017towards} as a basis to improve the automatic 
example sentence selection task which aims to select
sentences that clarify the differences between confusing words. 
GiveMeExample proposes a clarification score to represent the
ability of a sentence to clear up confusion between the given words. In this section, we 
describe the three main steps to build the automatic example sentence
selection model: the definition of the clarification score, the word usage
model, and the dictionary-like sentence classifier.

\subsection{Problem Definition}
Here we define the task more clearly. Given a confusing word set $W
= \{w_1, w_2, ..., w_n\}$ and their corresponding sentence sets $\{S_1, S_2,
..., S_n\}$, each sentence set contains a set of sentences $S_t = \{s_{t1},
s_{t2}, ..., s_{tm}\}$. The target is to choose $k$ sentences from each sentence
set that clarify the differences among the words in the confusing word set. The
desired results are thus sentence sets which clarify $W$, $\{S'_1, S'_2, ..., S'_n\}$,
where $S'_t = \{s'_1, s'_2, ..., s'_k\}$. 

\subsection{Workflow}
Given a word set and the corresponding sentence sets, GiveMeExample selects
sentences by (1) building a word usage model for each word, (2)
selecting learning-suitable sentences using a dictionary-like sentence classifier,
and (3) ranking sentences by computing clarification scores with the help of the
word usage model. The top five sentences for each word are selected to show 
learners. 

\subsection{Clarification Score}
To understand the definition of clarification, we start from the
confusing word set \{\textit{refuse}, \textit{reject}\} in
Table~\ref{tab:clarification_example}. The first sentence 
clarifies the differences better than the second sentence, as the usage
of \textit{refuse} in \say{refused his request} from the second sentence is the same
as that for \textit{reject} in \say{rejected his request} in the third
sentence. This illustrates two properties of clarification: the fitness
score and the relative closeness score. The fitness score measures how well
a sentence $s$ illustrates the usage of word $w_1$: in this sentence the word should 
be used in a common way instead of a rare way.
The relative closeness score, in turn, measures how well a sentence
$s$ for word $w_1$ highlights the difference between $w_1$ and the other words
$\{w_2, ... ,w_n\}$: it must be appropriate for $w_1$ but inappropriate for
$\{w_2, ..., w_n\}$. Namely, when we replace $w_1$ with $\{w_2, ..., w_3\}$ in
$s$, this sentence should become a wrong sentence. As a result, given a
function $P(s|w)$ that estimates the fitness between a sentence $s$ and a
word $w$, we define the clarification score as
\begin{equation}
    \small
    score(s|w_i) = P(s|w_i)*(\sum_{w_j \in W-w_i} P(s|w_i) - P(s|w_j))
    \label{eq:clarifun}
\end{equation}
which is the multiplication of the fitness score and the relative closeness score. 

\subsection{Word Usage Model}
The word usage model represents the distribution of the usage and
the context for a given word, that is, the fitness score $P(s|w)$.
GiveMeExample includes two word usage models: a Gaussian mixture
model (GMM) and a bidirectional long-short-term-memory model (BiLSTM),
described as follows. Notice that the word usage model is trained as a classifier
per word. 

\subsubsection{GMM with Local Contextual Features}
The idea of the GMM is to turn words around the target word, namely, its
context, into embeddings and then model the distribution with a Gaussian mixture
model~\cite{xu1996convergence}. 
Empirically, taking words within a window of size two provides the best results.
Therefore, given a sentence $s = \{w_1 \cdots w_t \cdots w_n\}$ where $w_t$ is
the target word, the features are $f=\{e_{w_{t-2}}, e_{w_{t-1}},
e_{w_{t-2}}+e_{w_{t-1}}, e_{w_{t+1}}, e_{w_{t+2}}, e_{w_{t+1}}+ e_{w_{t+2}}\}$.
Note that the features contain not only the corresponding word embeddings,
but also the summation of two adjacent words to leverage the meaning. 
Since the word embedding contains both word identity information and
semantic information, the GMM model\footnote{Each GMM model is trained on 5,000
instances.} therefore learns the distribution of both usage and semantic
meaning. 

\subsubsection{BiLSTM}
As the confusing words can diverge widely from the target word
itself, 
or could involve long-term dependencies,   
GMM with local
contextual features do not always capture enough information. 
The BiLSTM model thus utilizes the whole sentence as a feature. 
The BiLSTM model consists of a forward LSTM and a backward LSTM,
which take the words preceding and following the target word as features
respectively. 
The output vectors of these two LSTMs are concatenated to form a sentence
embedding. After passing through two dense layers, the BiLSTM model is then built as a
binary classifier that decides whether the given sentence is the sentence of the
target word or not. In contrast to the generative GMM model, negative samples
are needed to train the BiLSTM. As a result, sentences from the corpus are
randomly sampled as negative samples\footnote{Each BiLSTM model is trained on 5,000
positive instances and 50,000 negative instances.}.

\subsection{Dictionary-like Sentence Classification}
The given sentences are not always suitable for language learning. For example,
a 40-word-long sentence could be too complicated and distracting to learn, and a
short sentence such as \say{It is sophisticated} is not suitable for language learning
due to its lack of information.
GiveMeExample is equipped with a dictionary-like sentence classifier to select
sentences that are simple but informative. GiveMeExample collects sentences
from the COBUILD English Usage Dictionary~\cite{sinclair1992collins} to train the
dictionary-like sentence classifier with syntactic 
features~\cite{pilan2014rule} and a logistic regression model~\cite{walker1967estimation}.
Hence, it tends to select sentences similar to those in the COBUILD dictionary.

\section{Deployment: Sentence Translation}





\begin{figure}
    \centering
    \includegraphics[width=0.8\columnwidth]{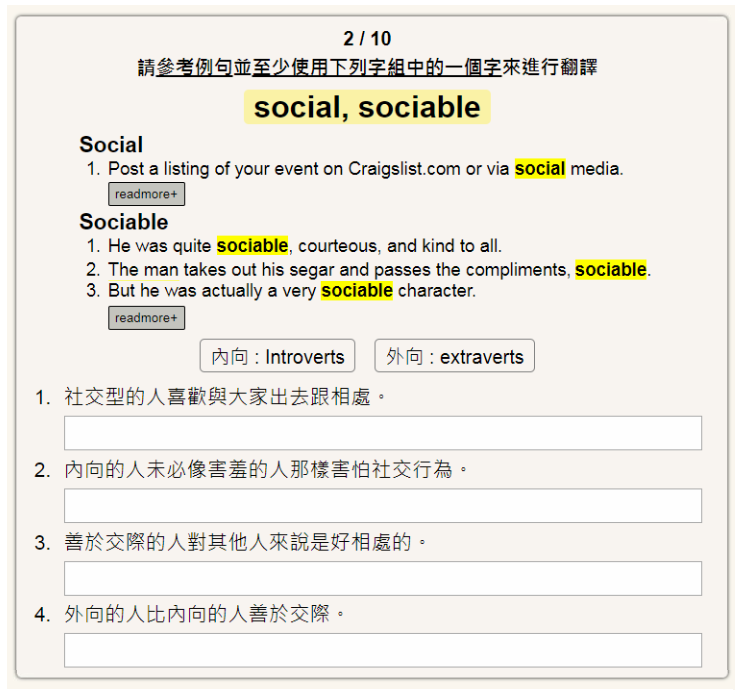}
	 \caption{Example questions for translation experiment. Participants 
	 click the readmore button to retrieve more example sentences (the maximum number
	 of sentences for each word is five). Also, \textit{introverts} and \textit{extraverts} are
	 two tips that we provide, as they are more difficult but not directly related to
	 \textit{social} and \textit{sociable}. }
	 \vspace{-0.4cm}
    \label{fig:example_question}
\end{figure}

\input{tables/scoring_example.tex}

The sentence translation experiment was separated into a pre-test and
a post-test. In both of the tests, participants were asked to translate ten sets of
questions from Mandarin to English. In each set, there were four translation
questions corresponding to a specific set of confusing words. In addition to
answering the question, participants could refer to the example sentences
suggested by GiveMeExample in the post-test. In the following paragraph, we
describe the experiment in detail.

\subsection{Building Translation Questions}
In the sentence translation task were 15 confusing word sets selected from
Collins COBUILD English Usage~\cite{sinclair1992collins} and the Longman Dictionary
of Common Errors~\cite{turton1996longman}. These two books identify errors in word 
usage commonly made by language learners and then clear up the confusion. Thus
the word sets provided in the books were used as the desired confusing words.
A word set contained two or three words. After selecting the confusing word set,
we extracted sentences that contain these words from the parallel corpora
Chinese English News Magazine Parallel Text (LDC2005T10) and Hong Kong Parallel
Text (LDC2004T08). These sentences were used as candidate questions. Since many
sentences in the parallel corpora were long and complicated, we removed
sentences whose Chinese translation contains more than 40 words. In the last step,
we manually chose appropriate sentences for testing the confusing words. In
the end, 15 confusing word sets were determined, each of which contains four
questions to be translated resulting a total of 60 questions. Note that some difficult words in the
question, such as \say{introverts} and \say{extraverts} in Figure~\ref{fig:example_question}, were provided as they were unrelated to testing learner use of confusing words.

\subsection{Recommending Example Sentences}
To recommend sentences, we first collected sentences from Vocabulary\footnote{https://www.vocabulary.com/},
an online dictionary. The example
sentences in Vocabulary mainly come from formally-written news articles.
We collected 5,000 sentences for each word and used all of them to
train the GMM and BiLSTM word usage models. When recommending example sentences,
we used only the qualified sentences which were filtered by the dictionary-like
sentence classifier. The pretrained 300-dimension
GloVe~\cite{pennington2014glove} embeddings were used in both GMM and BiLSTM. 
We selected the last five sentences from Vocabulary as a baseline setting.

\subsection{Experimental Setup}
Sixteen college students were recruited for this translation experiment. As the translation of total 60 questions may not be done in one class, each
participants was asked to complete ten randomly-assigned question sets, each of which
contained four questions. Thus a total of 40 translation
questions were given. This process guarantees that every questions is translated by the same number of participants. The testing period was about 45 minutes, leaving
participants about five minutes for each question set. In addition to
translating, five example sentences were provided for each word in the post-test.
To ensure the students read the suggested sentences, only one example sentence was displayed in the beginning, a \say{readmore} button was designed for retrieving more example sentences  
(the maximum number of example sentences is five for each word). 
The \say{readmore} activities were logged for further investigation.
The pre- and post-tests were administered in two different weeks to reduce
short-term memory effects. Figure~\ref{fig:example_question} shows a
screen-shot of a post-test with the confusing word set \textit{social} and
\textit{sociable}. 

The example sentences provided were suggested by the GMM and
BiLSTM models or selected from the Vocabulary website. Note that to discourage
participants from guessing specific patterns, the example sentences from 
one of the three sources were presented randomly. For instance, as
GMM takes contexts within a window as features, the most significant difference
exists only within this window. However, we do not expect participants to look only
at this small piece of text. Also, sentences from Vocabulary are generally
more difficult than those from GMM or BiLSTM, but participants who are consistently
presented with difficult sentences may stop considering these example 
sentences to be useful resources. As the source is assigned randomly for each proposed example sentence,
the total number of sentences for each source is set to the number that can best distribute sentences from different sources evenly. 

\subsection{Grading}
Grading was done by an English native speaker who is professional in language
learning and teaching. The grading criteria takes into account appropriateness, 
grammar, and completeness. Appropriateness measures whether the
correct word is used or not, so the score here is either zero or one point.
Grammar involves local grammar as well as global grammar. All the grammar
errors relating directly to the target confusing word belong to local grammar;
the remaining grammar errors throughout the sentence belong to global
grammar. The initial points for both grammar parts are four points; each
grammar error results in a one-point deduction. Completeness, which evaluates
whether the student's translation represents all of the meanings,
takes into account structure and meaning. If a student
missed content such as adverbial phrases, points were deducted in terms of
structure. Similarly, if a student's translation was different from the
original meaning, points were deducted in terms of meaning. Both structure
and meaning started with two points. Examples are listed in
Table~\ref{tab:scoring_example}. Given our focus on examining whether
students can learn how to differentiate and use confusing words, we
computed a weighted sum for reference as follows:
\begin{equation}
    \footnotesize
    Weighted Sum = 5*Appropriateness + Local Grammar
\end{equation}
which is the sum of the appropriateness scores, weighted by 5, and the 
local grammar scores.


\input{tables/result_high_low.tex}

\section{Results and Discussions}

The pre and post scores for the grading categories are summarized in
Table~\ref{tab:result_high_low}. 
Student are separated into Highly proficient group and Less proficient group evenly by an external collocation test score \cite{chen2011factors}.
In general, the suggested example sentences 
helped students make substantial progress in terms of sentence structure. It is
worth noting that students were able to comprehend the meaning of confusing words in
the given sentences selected from both of the BiLSTM and GMM models. Students
performed significantly better in \textit{appropriateness}, \textit{local grammar},
and \textit{structure} when the sentences were suggested by BiLSTM; while 
the GMM model was good at presenting the structures of sentences and
demonstrating the meaning of confusing words. 

Highly proficient students
learned confusing words better from the suggested example sentences. The
findings showed that BiLSTM helped them gain a better understanding of
\textit{appropriateness}, \textit{local grammar}, and \textit{structure}, and GMM
helped with \textit{structure} and \textit{meaning}. Although
it was difficult for less-proficient students to recognize the difference 
(small improvement in \textit{appropriateness} and \textit{local grammar}), the
GMM model significantly facilitated their comprehension in terms of 
\textit{structure}, \textit{global grammar} and \textit{meaning}.

The \say{readmore} logs show that most of the students clicked the button and expand all the example sentences immediately. 
This might imply that students did read all the example sentences and could refer to them when producing translations.


We analyzed the translation tasks to identify possible problems.
Below we discuss three possible explanations in terms of test items,
learner behavior, and the suggested example sentences. 

First, in the proposed translation task, participants sometimes focused on the wrong
segment of the test item to translate with the confusing words. This may be
because in this productive testing process, we do not specifically
tell participants which source word should be aligned to the target confusing
word. For instance, in \say{For a person to become so poor, if it's not
because they didn't work hard in their youth then it’s because they have truly
had hard luck}, participants should have translated the
source words \say{hard luck} to English using the appropriate word in the
confusing word set. However, the students showed confusion in their focusing on
translating the source word \textit{poor} into one of \textit{hard},
\textit{difficult}, and \textit{tough} as opposed to the source word \textit{hard}
in \textit{hard luck}. One example translation made by a participant is
\say{The reason why a person's life is tough might because he/she was lazy when
he/she was young or he/she had a bad luck}. In such cases, the learning effect
cannot be correctly evaluated.

We seek to find the best example sentences for word sets where the words are confusing
for learners. Hence regarding the suggested example sentences, the example
sentences were extracted as long as the confusing words shared the least
familiar senses. However, this led to words being chosen in example sentences
with different senses and/or even different parts of speech,
which is how we wanted to compare them. The words
\textit{hard} and \textit{difficult} exemplify these issues. First, according to
WordNet, \textit{hard} in this case indicates \say{resisting weight or
pressure} in the example sentence \say{Such uncertainty can be hard on
families, too}, whereas \textit{difficult} means \say{needing skill or effort}
in the sentence \say{But other stories are more difficult to
explain}. On the other hand, \textit{hard} is an adverb in \say{Banks will have
to work harder to make profits}, while \textit{difficult} is an adjective in
\say{But other stories are more difficult to explain}.

Student behavior also affected the performance of
this study. Some highly proficient students were observed skipping the example
sentences and thus not learning from them how to differentiate the confusing words, which
led to inappropriate translations similar to those made in the pre-test. It
could be that these highly proficient students were more confident of
their command of certain confusing words. For example, when required to choose
from \textit{beat}, \textit{defeat}, and \textit{win} to translate \say{Emmanuel
Macron beats Marine Le Pen in both rounds of the French presidential
election}, one highly proficient student made these translations in the pre- and
post-tests, respectively: \say{Emmanuel Macron won over Marine Le Pen for two rounds of
presidential election}, and \say{Emmanuel Macron won over Marine Le Pen for
presidential election for two rounds}, whereas \textit{win over} is not a usage
suggested by example sentences. In addition, from this example we can see that though they rarely read example sentences, they did try to translate in other words in the post-test, which results in the unstable scores of \textit{global grammar} that are less relevant to the near-synonym recognition but to the translation instead.

These three limitations partially explain learner performance in the
translation task.
Thus we attempted to refine the method for example sentence extraction.
Improving the test items and controlling student learning behavior is
beyond the scope of this study.

\section{Leveraging First Language for Better Example Sentence Selection}
From the results of the translation experiment, we observed that some words were
confusing to students due to language transfer from L1 (native language)
to L2 (foreign language). Some students learn English such that they only
remember how to spell words and their L1 definitional glosses, rather than
understanding their context or usage. For example, the confusing words
\textit{hard} and \textit{difficult} are very similar and almost
interchangeable. If these words are memorized only by memorizing the L1 definitional
glosses, \textit{not easy}, students may fail to recognize the slight difference
between them. 
In other words, example sentences containing words that translate into similar
glosses in L1 are the sentences that indeed contain confusing senses, and thus
are the target candidates for the GiveMeExample system to consider for suggestion.
We follow this line of thinking to improve the example sentences.
 
In the new setting, the GiveMeExample system groups example sentences by the L1
definitional glosses of confusing words before proceeding to automatic
sentence selection with the BiLSTM or GMM word usage model. When a word has
multiple senses, this step helps to identify the confusing sense, under the
assumption that words with similar L1 definitions are confusing. Take for example
\textit{hard} and \textit{difficult}: \textit{hard} as an
adjective has multiple meanings~-- ``not easy, requiring great physical or mental
effort to accomplish, resisting weight or pressure, hard to bear'', etc;
whereas \textit{difficult} has the meanings ``not easy, requiring great physical or
mental effort to accomplish, and hard to control''. The common sense in L1 is
\textit{not easy, requiring great physical or mental effort to accomplish.}
Sentences containing confusing words whose L1 translations share these two
senses are selected for later processing and suggestion.

To identify these sentences, we need each word in the sentence and its
corresponding L1 translation. For this purpose, parallel texts from two
corpora~-- Chinese English News Magazine Parallel Text (LDC2005T10) and Hong Kong
Parallel Text (LDC2004T08), that is, a total of 2,682,129 English-Chinese
sentence pairs~-- are utilized to learn the word alignment between L1 and L2
parallel sentences.
To align example sentences from Vocabulary, first they were all translated into Traditional Chinese using Googletrans\footnote{https://pypi.org/project/googletrans/}. 
Then we used NLTK\footnote{https://www.nltk.org/} to tokenize English sentences 
and CKIP~\cite{chen1992ckip} to segment Chinese sentences respectively. After that, the
word alignment model GIZA++~\cite{och03:asc}, a toolkit that implements
several statistical word alignment models, was adopted to align English words
to their corresponding Chinese words.
After alignment, the L1 translations of confusing words were 
recognized,   
after which the sentences in the example sentence pool of the confusing words in the same
set were clustered with respect to their L1 translation. There were 12 confusing
word sets with more than one common L1 translation. Only words in three
confusing word sets (\textit{possibility} vs. \textit{opportunity},
\textit{social} vs. \textit{sociable}, and \textit{unusual} vs.
\textit{strange}) had all different L1 translations. When a common L1
translation was found for a set of confusing words, GiveMeExample passed through
only those sentences containing confusing words with the same L1 translation to the
sentence selection component.

\subsection{Human Evaluation}
We employed Amazon Mechanical Turk crowd-workers to give their perspectives on
the suggested sentences considering the L1 of learners. Twelve sets of confusing words
with common L1 translations were evaluated. GiveMeExample in both the
original and the new settings suggested respectively five sentences using the
BiLSTM and GMM models for each word in the twelve sets. In this new setting, six
words~-- (\textit{briefly}, \textit{duty}, \textit{ordinary}, \textit{sight},
\textit{shortly}, and \textit{unusual})~-- had less than five sentences.

\begin{figure}
    \centering
    \includegraphics[width=1.0\columnwidth]{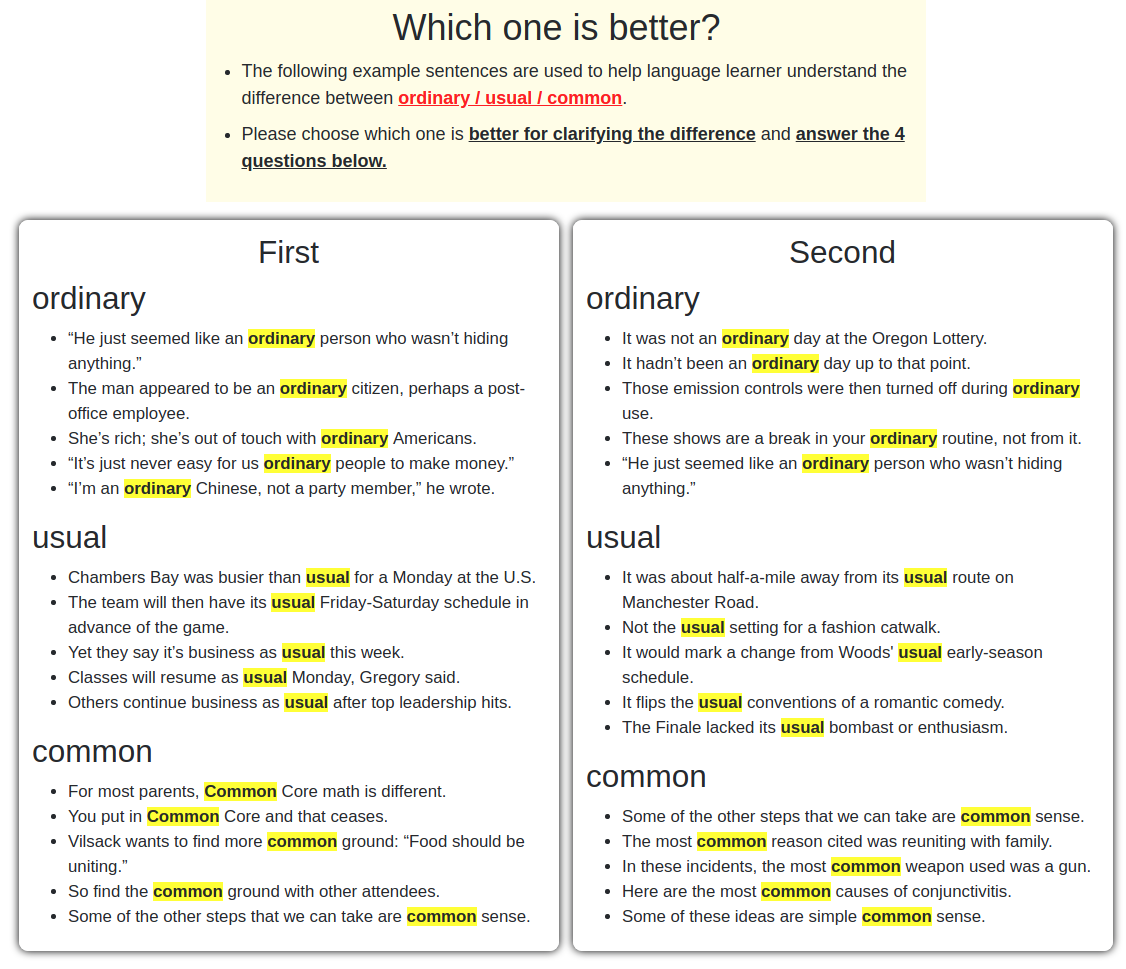}
	 \caption{An example survey for crowd-workers to compare GiveMeExample with
	 different settings. In this specific example, \textit{first} represents 
	 sentences suggested by the new setting; \textit{second} represents those 
	 from the original.}
	 \vspace{-0.4cm}
    \label{fig:turkers' experiment}
\end{figure}

Figure~\ref{fig:turkers' experiment} shows a screenshot of two versions of the
suggested example sentences presented side-by-side. Crowd-workers were given no
information about the settings or the sentence selection models (BiLSTM or GMM).
For each task, participants were to read several sentences suggested by the two
versions of the GiveMeExample system and then answer the
following four questions.

\begin{enumerate}[label=Q\arabic*:]
\item Is Mandarin your first language (y/n)?  \vspace{-0.2cm}
\item Are these words confusing to you (y/n)?  \vspace{-0.2cm}
\item Which set of example sentences you think is more useful for learning these words (1/2)?  \vspace{-0.2cm}
\item In what aspect you think they are more useful (choose one)? 
(a) clarifying their meaning (e.g., \textit{\textbf{social} encounter} vs.
\textit{\textbf{sociable} character})
(b) demonstrate their usage (e.g., \textit{as \textbf{usual}} but not
\textit{as \textbf{common}})
(c) showing correct grammar (e.g., \textit{The proposal was narrowly
\textbf{defeated} in a January election}, but \textit{Obviously we want to
continue to \textbf{win} games}.)
\end{enumerate}

The purpose of Q1 and Q2 is to understand the background of turkers, Q3 is to compare the new setting with old setting among two models, and Q4 is to investigate the effect of considering L1 translation.
We also consulted a native speaker who works as an expert editor. This expert
completed the surveys under the same conditions as the crowd-workers. 

\input{tables/survey_result.tex}

\subsection{Results and Analysis}
Sixty-one crowd-workers participated in the evaluation. Mandarin was the first 
language of 12 (19.67\%) of them. On average, each worker completed six
tasks (SD=8.17). For each set out of 12 sets, 15 workers were asked to answer
the questions. We tested the example sentences suggested by both GMM and BiLSTM
models, collecting in total 360 ratings from workers. It was an interesting finding that only 5\% of the
confusing word sets were labeled by workers as confusing no matter they were native speakers or not \footnote{The expert
had a clear understanding of these words.}. Details are shown in
Table~\ref{tab:survey_result}.

Table~\ref{tab:survey_result} shows the feedback on Q3 and Q4 from workers and the expert on each confusing word set. Results from the expert
confirm that when considering L1, our approach could provide better example
sentences. However, results from the crowd-workers were mixed. 

Several interesting observations were gleaned from this experiment.
First, when considering the L1 translation and grouping sentences by their L1
sense, the example sentences containing confusing words with different senses
were excluded. Therefore, learners could focus more on the confusing sense to be 
learned. For example, \textit{work hard} is a commonly seen phrase in the
example sentences suggested by the original setting. When students learned the
confusion set containing \textit{hard}, \textit{difficult}, and \textit{tough},
the sentences containing \textit{work hard} were of little help, as the meanings
were irrelevant to the confusing sense in this set. However, in the new
setting, the example sentences for \textit{hard} were more semantically related to
\textit{difficult} and \textit{tough}. 
We can say that in this task, consideration of L1 amounted to implicitly performing word sense disambiguation (WSD).  


The exclusion of sentences that did not contain words with the confusing sense
has additional benefit. That is, the suggested sentences are more likely to focus on
the demonstration of the confusing sense. This has the advantage that
the confusing words in the suggested sentences are diverse in their part of
speech and pragmatic domain.
For instance, in the confusion set \textit{defeat}, \textit{win}, and \textit{beat}, the common L1 sense among them is \say{to conquest} and \say{victory}. Under these certain meanings, only \textit{win} can be used as a verb or a noun whereas the other two words can only function as a verb. This illustrates the power of grouping sentences by L1 translation
Another example is
\textit{destroy} in the confusion set \textit{destroy}, \textit{ruin}, and
\textit{spoil}.
In the original setting, \textit{destroy} is used in only the military domain
and thus is misleading. When using the GMM model which considers only the local
context, the issue is even more serious. This is mitigated in the new setting,
especially for the GMM model.


Following the above, in some cases workers indeed tended to prefer example
sentences of some pattern. For example, in the set \textit{scarce},
\textit{rare}, and \textit{unusual}, confusing words in the example sentences
that shared the L1 translation \textit{very hardly} resulted in example sentences
containing confusing words functioning as adverb, adjective, and adjective,
respectively; however, in the original setting where 
context is considered before sense,  
they all function as adjectives. This interesting result
reveals that there is overhead when learning from materials without patterns,
which could also be why only highly proficient students can learn
the appropriateness. 

\section{Conclusion}
In this paper, we leverage GiveMeExample, an AI system which automatically
suggests example sentences to help ESL learners better learn to differentiate
confusing words. To evaluate the system effectiveness, we designed a sophisticated
sentence translation task around the problem of students
not really learning via the previously designed receptive task, i.e.,
multiple-choice selection. This approach was evaluated using college students;
results show that students made substantial progress with assistance of the
system. Specifically, after learning the example sentences, students produced
more structural sentences. However, learning to use appropriate words is a
demanding task which requires higher language proficiency.

The learner's first language may lead to confusion in different areas: this is
also taken into account with a novel approach. Overall, the example sentences in the
refined list were considered more useful for learning by Amazon mechanical
turkers and the expert English editor. However, for ESL learners such as
students and some of the turkers, they tended to prefer example sentences with
similar patterns to mitigate cognitive overhead. Thus, future work will
focus on providing example sentences with similar patterns but diverse contexts.

\section*{Acknowledgments}
This research is partially supported by Ministry of Science and Technology, Taiwan under Grant No. MOST108-2634-F-002-008- and  MOST108-2634-F-001-004-.


\bibliography{acl2019}
\bibliographystyle{acl_natbib}


\end{document}

%% file: tables/clarification_example.tex
\begin{table*}[t]
    \centering
    \footnotesize
    \begin{tabular}{cll}
        \toprule \hline
        \multicolumn{1}{c}{\textbf{Number}} & \multicolumn{1}{c}{\textbf{Word}} & \multicolumn{1}{c}{\textbf{Example sentence}}\\ \hline
        1 & refuse & I was expecting you to \underline{refuse} to leave the house.\\
        2 & refuse &  She declined to serve as an informant and \underline{refused} his request that she keep their meeting secret. \\
        3 & reject & In July, a judge in Australia \underline{rejected} his request for a suppression order. \\ \hline \bottomrule
    \end{tabular}
    \caption{Example sentences that illustrate clarification}
    \vspace{-0.4cm}
    \label{tab:clarification_example}
    
\end{table*}

%% file: tables/scoring_example.tex
\begin{table*}[t]
    \centering
    \footnotesize
    \begin{tabular}{llc}
        \toprule \midrule
        \multicolumn{1}{c}{\textbf{Category}} & \multicolumn{1}{c}{\textbf{Example}} & \multicolumn{1}{c}{\textbf{Grade}} \\ \midrule 
         \textbf{Appropriateness} &  There is a small \sout{opportunity} \underline{\textbf{possibility}} that she had actually met such a person. & 0\\ \midrule
         \textbf{Local grammar} & What are you going to do if we \underline{refuse} to \sout{following} \textbf{follow} you? & 3 \\\midrule
         \textbf{Global grammar} & This building \sout{is} \textbf{was} \underline{destroyed} by the earthquake. & 3 \\\midrule
         \multirow{2}{*}{\textbf{Structure}} & The accident was caused \sout{by \underline{error}}. & \multirow{2}{*}{1} \\
         & (The error is made by human, so it should be \say{by human error.}) & \\\midrule
         \multirow{3}{*}{\textbf{Meaning}} & To a \underline{skillful} pilot, \sout{it's lucky to say that landing in torrential rain.} & \multirow{3}{*}{1}\\ 
         & (The meaning is wired and the correct sentence should be \say{Landing safely in torrential & \\ 
         & rain can only be a matter of luck for the most skilled pilot.}) & \\ \midrule \bottomrule
    \end{tabular}
    \caption{Examples of grade criteria. The underlined word is the target confusing word.}
    \vspace{-0.4cm}
    \label{tab:scoring_example}
\end{table*}

%% file: tables/result_high_low.tex
\begin{table*}[t]
    \centering
    \scriptsize
    \addtolength{\tabcolsep}{-0.125cm} 
    \begin{tabular}{cc|ccc|ccc|ccc|ccc|ccc|ccc}
        \toprule \hline
        \multirow{2}{*}{Group} & \multirow{2}{*}{Model} & \multicolumn{3}{c|}{Appropriateness} & \multicolumn{3}{c}{Local grammar} & \multicolumn{3}{|c}{Weighted sum} & \multicolumn{3}{|c|}{Global grammar} & \multicolumn{3}{c}{Structure} & \multicolumn{3}{|c}{Meaning}\\ 
        
         & & pre & post & t-test & pre & post & t-test & pre & post & t-test & pre & post & t-test & pre & post & t-test & pre & post & t-test \\ \hline
        H & Vocabulary & \textbf{0.714} & 0.571 & 0.302 & \textbf{3.429} & 3.143 & 0.178 & \textbf{7.000} & 6.000 & 0.237 & \textbf{2.429} & 2.143 & 0.229 & 0.714 & \textbf{1.000} & 0.229 & 1.000 & 1.000 & 0.500 \\ 
        H & GMM & 0.444 & 0.444 & 0.500 & 2.444 & \textbf{3.000} & 0.123 & 4.667 & \textbf{5.222} & 0.347 & \textbf{1.667} & 1.556 & 0.364 & 0.667 & \textbf{1.222} & 0.025* & 0.333 & \textbf{1.111} & 0.004*\\ 
        H & BiLSTM & 0.273 & \textbf{0.545} & 0.041* & 2.364 & \textbf{3.273} & 0.008* & 3.727 & \textbf{6.000} & 0.011* & \textbf{1.545} & 1.364 & 0.276 & 0.818 & \textbf{1.182} & 0.052 & 0.545 & \textbf{0.909} & 0.052\\ \hline
        L & Vocabulary & 0.182 & \textbf{0.364} & 0.170 & 2.182 & \textbf{2.909} & 0.098 & 3.091 & \textbf{4.727} & 0.056 & 0.818 & \textbf{1.091} & 0.247 & 0.364 & \textbf{1.000} & 0.013* & 0.455 & \textbf{0.636} & 0.220\\
        L & GMM & 0.417 & \textbf{0.583} & 0.169 & 2.333 & \textbf{2.917} & 0.066 & 4.417 & \textbf{5.833} & 0.072 & 0.750 & \textbf{1.500} & 0.028* & 0.500 & \textbf{1.167} & 0.012* & 0.333 & \textbf{1.083} & 0.010*\\
        L & BiLSTM & 0.429 & \textbf{0.524} & 0.165 & 2.667 & \textbf{2.714} & 0.443 & 4.810 & \textbf{5.333} & 0.169 & 1.238 & \textbf{1.571} & 0.116 & 0.762 & \textbf{1.143} & 0.004* & 0.524 & \textbf{0.857} & 0.025*\\ \hline \bottomrule
        
        

    \end{tabular}
    \addtolength{\tabcolsep}{0.125cm} 
	 \caption{Result of translation experiment. The number of translated questions for each model ranges from 7 to 21, with the average number 11.8, depending on the number of early leave and absence we encountered in the experiment day. The pre- and post- numbers 
	 correspond to the average score for pre-test and post-test respectively and the
	 t-test stars represent significance. The participants were separated
	 into highly proficient (H) and less proficient (L) groups.}
	 \vspace{-0.4cm}
    \label{tab:result_high_low}
\end{table*}

%% file: tables/survey_result.tex
\begin{table*}[]
\centering
\scriptsize
\addtolength{\tabcolsep}{-0.10cm}
\begin{tabular}{ccccccccccccc}
\toprule \midrule
\multirow{3}{*}{Confusing word set} & \multicolumn{2}{c}{Q2 (turkers)} & \multicolumn{2}{c}{Q3 (turkers)} & \multicolumn{2}{c}{Q3 (expert)} & \multicolumn{6}{c}{Q4 (turkers)} \\ \cline{2-13} 
 & \multirow{2}{*}{No} & \multirow{2}{*}{Yes} & \multirow{2}{*}{BiLSTM} & \multirow{2}{*}{GMM} & \multirow{2}{*}{BiLSTM} & \multirow{2}{*}{GMM} & \multicolumn{3}{c}{BiLSTM} & \multicolumn{3}{c}{GMM} \\ \cline{8-13} 
 &  &  &  &  &  &  & (a) & (b) & (c) & (a) & (b) & (c) \\ \hline
ordinary / usual / common & 97\% & 3\% & N & N & O & N & 46\% & 40\% & 14\% & 46\% & 40\% & 14\% \\
skillful / skilled & 93\% & 7\% & O & O & O & N & 46\% & 46\% & 6\% & 34\% & 46\% & 20\% \\
alternative / alternate & 97\% & 3\% & O & O & N & O & 34\% & 60\% & 6\% & 26\% & 46\% & 26\% \\
destroy / ruin / spoil & 100\% & 0\% & O & N & N & N & 40\% & 40\% & 20\% & 20\% & 74\% & 6\% \\
scarce / rare / unusual & 97\% & 3\% & O & O & O & O & 14\% & 74\% & 14\% & 0\% & 74\% & 26\% \\
defeat / win / beat & 100\% & 0\% & N & N & N & N & 40\% & 46\% & 14\% & 20\% & 60\% & 20\% \\
sight / landscape / scenery & 93\% & 7\% & N & O & N & O & 40\% & 46\% & 14\% & 34\% & 46\% & 20\% \\
briefly / shortly / concisely & 97\% & 3\% & O & N & O & O & 14\% & 66\% & 20\% & 14\% & 60\% & 26\% \\
hard / difficult / tough & 90\% & 10\% & O & N & O & O & 14\% & 80\% & 6\% & 20\% & 54\% & 26\% \\
error / mistake / oversight & 90\% & 10\% & O & N & N & N & 26\% & 60\% & 14\% & 20\% & 66\% & 14\% \\
duty / job / task & 97\% & 3\% & N & N & O & N & 46\% & 46\% & 6\% & 14\% & 66\% & 20\% \\
obligation / responsibility / commitment & 93\% & 7\% & N & N & N & N & 26\% & 66\% & 6\% & 46\% & 40\% & 14\% \\ \hline
\multicolumn{1}{r}{Mean} & 95\% & 5\% & 42\%(N) & 67\%(N) & 50\%(N) & 58\%(N) & 32\% & 56\% & 12\% & 24\% & 56\% & 20\% \\ 
\midrule
\toprule
\end{tabular}%
\addtolength{\tabcolsep}{0.10cm}
\caption{Results from the human evaluation. N represents the example
sentences from the new setting, and O are from the original one. In addition,
the expert annotated that ALL of the suggested sentences were useful for
demonstrating their usage (b).}
\vspace{-0.4cm}
\label{tab:survey_result}
\end{table*}

%% file: acl2019_main.bbl
\begin{thebibliography}{41}
\expandafter\ifx\csname natexlab\endcsname\relax\def\natexlab#1{#1}\fi

\bibitem[{Alikaniotis et~al.(2016)Alikaniotis, Yannakoudakis, and
  Rei}]{alikaniotis2016automatic}
Dimitrios Alikaniotis, Helen Yannakoudakis, and Marek Rei. 2016.
\newblock Automatic text scoring using neural networks.
\newblock \emph{arXiv preprint arXiv:1606.04289}.

\bibitem[{Alonso~Ramos et~al.(2014)Alonso~Ramos, Garc{\'\i}a~Salido, and
  Vincze}]{alonso2014towards}
Margarita Alonso~Ramos, Marcos Garc{\'\i}a~Salido, and Orsolya Vincze. 2014.
\newblock Towards a collocation writing assistant for learners of spanish.

\bibitem[{Chen and Liu(1992)}]{chen1992ckip}
Keh-Jiann Chen and Shing-Huan Liu. 1992.
\newblock Word identification for {M}andarin {C}hinese sentences.
\newblock In \emph{Proceedings of the 14th Conference on Computational
  linguistics (COLING '92) - Volume 1}, pages 101--107.

\bibitem[{Chen and Lin(2011)}]{chen2011factors}
Mei-Hua Chen and Maosung Lin. 2011.
\newblock Factors and analysis of common miscollocations of college students in
  {T}aiwan.
\newblock \emph{Studies in English Language and Literature}, (28):57--72.

\bibitem[{Dong and Zhang(2016)}]{dong2016automatic}
Fei Dong and Yue Zhang. 2016.
\newblock Automatic features for essay scoring--an empirical study.
\newblock In \emph{Proceedings of the 2016 Conference on Empirical Methods in
  Natural Language Processing}, pages 1072--1077.

\bibitem[{Doshi-Velez and Kim(2017)}]{doshi2017towards}
Finale Doshi-Velez and Been Kim. 2017.
\newblock Towards a rigorous science of interpretable machine learning.
\newblock \emph{arXiv preprint arXiv:1702.08608}.

\bibitem[{Edmonds and Hirst(2002)}]{edmonds2002near}
Philip Edmonds and Graeme Hirst. 2002.
\newblock Near-synonymy and lexical choice.
\newblock \emph{Computational linguistics}, 28(2):105--144.

\bibitem[{Futagi(2010)}]{futagi2010effects}
Yoko Futagi. 2010.
\newblock The effects of learner errors on the development of a collocation
  detection tool.
\newblock In \emph{Proceedings of the fourth workshop on Analytics for noisy
  unstructured text data}, pages 27--34. ACM.

\bibitem[{Hashemi and Gowdasiaei(2005)}]{hashemi2005attribute}
Mohammad~Reza Hashemi and Farah Gowdasiaei. 2005.
\newblock An attribute-treatment interaction study: {L}exical-set versus
  semantically-unrelated vocabulary instruction.
\newblock \emph{RELC journal}, 36(3):341--361.

\bibitem[{Hemchua et~al.(2006)Hemchua, Schmitt et~al.}]{hemchua2006analysis}
Saengchan Hemchua, Norbert Schmitt, et~al. 2006.
\newblock {An analysis of lexical errors in the English compositions of Thai
  learners}.

\bibitem[{Huang et~al.(2017)Huang, Chen, and Ku}]{huang2017towards}
Chieh-Yang Huang, Mei-Hua Chen, and Lun-Wei Ku. 2017.
\newblock Towards a better learning of near-synonyms: {A}utomatically
  suggesting example sentences via fill in the blank.
\newblock In \emph{Proceedings of the 26th International Conference on World
  Wide Web Companion}, pages 293--302. International World Wide Web Conferences
  Steering Committee.

\bibitem[{Karlsson(2014)}]{karlsson2014advanced}
Monica Karlsson. 2014.
\newblock {Advanced Students' L1 and L2 Mastery of Lexical Fields of Near
  Synonyms}.
\newblock \emph{World Journal of English Language}, 4(3):1.

\bibitem[{Laufer(1990)}]{laufer1990ease}
Batia Laufer. 1990.
\newblock Ease and difficulty in vocabulary learning: {S}ome teaching
  implications.
\newblock \emph{Foreign Language Annals}, 23(2):147--155.

\bibitem[{Laufer and Girsai(2008)}]{laufer2008form}
Batia Laufer and Nany Girsai. 2008.
\newblock Form-focused instruction in second language vocabulary learning: A
  case for contrastive analysis and translation.
\newblock \emph{Applied linguistics}, 29(4):694--716.

\bibitem[{Liu and Zhong(2014)}]{liu2014l2}
Dilin Liu and Shouman Zhong. 2014.
\newblock {L2 vs. L1 use of synonymy: An empirical study of synonym
  use/acquisition}.
\newblock \emph{Applied linguistics}, 37(2):239--261.

\bibitem[{Martin(1984)}]{martin1984advanced}
Marilyn Martin. 1984.
\newblock Advanced vocabulary teaching: {T}he problem of synonyms.
\newblock \emph{The Modern Language Journal}, 68(2):130--137.

\bibitem[{McCarthy and Navigli(2007)}]{mccarthy-navigli-2007-semeval}
Diana McCarthy and Roberto Navigli. 2007.
\newblock \href {https://www.aclweb.org/anthology/S07-1009} {{S}em{E}val-2007
  task 10: {E}nglish lexical substitution task}.
\newblock In \emph{Proceedings of the Fourth International Workshop on Semantic
  Evaluations ({S}em{E}val-2007)}, pages 48--53, Prague, Czech Republic.
  Association for Computational Linguistics.

\bibitem[{McCarthy and Navigli(2009)}]{mccarthy2009english}
Diana McCarthy and Roberto Navigli. 2009.
\newblock The english lexical substitution task.
\newblock \emph{Language resources and evaluation}, 43(2):139--159.

\bibitem[{Melamud et~al.(2015)Melamud, Levy, and Dagan}]{melamud2015simple}
Oren Melamud, Omer Levy, and Ido Dagan. 2015.
\newblock A simple word embedding model for lexical substitution.
\newblock In \emph{Proceedings of the 1st Workshop on Vector Space Modeling for
  Natural Language Processing}, pages 1--7.

\bibitem[{Mihalcea et~al.(2010)Mihalcea, Sinha, and
  McCarthy}]{mihalcea2010semeval}
Rada Mihalcea, Ravi Sinha, and Diana McCarthy. 2010.
\newblock Semeval-2010 task 2: Cross-lingual lexical substitution.
\newblock In \emph{Proceedings of the 5th international workshop on semantic
  evaluation}, pages 9--14. Association for Computational Linguistics.

\bibitem[{Napoles and Callison-Burch(2017)}]{napoles2017systematically}
Courtney Napoles and Chris Callison-Burch. 2017.
\newblock Systematically adapting machine translation for grammatical error
  correction.
\newblock In \emph{Proceedings of the 12th Workshop on Innovative Use of NLP
  for Building Educational Applications}, pages 345--356.

\bibitem[{Ng et~al.(2014)Ng, Wu, Briscoe, Hadiwinoto, Susanto, and
  Bryant}]{ng2014conll}
Hwee~Tou Ng, Siew~Mei Wu, Ted Briscoe, Christian Hadiwinoto, Raymond~Hendy
  Susanto, and Christopher Bryant. 2014.
\newblock The {CoNLL-2014} shared task on grammatical error correction.
\newblock In \emph{Proceedings of the Eighteenth Conference on Computational
  Natural Language Learning: Shared Task}, pages 1--14.

\bibitem[{Och and Ney(2003)}]{och03:asc}
Franz~Josef Och and Hermann Ney. 2003.
\newblock A systematic comparison of various statistical alignment models.
\newblock \emph{Computational Linguistics}, 29(1):19--51.

\bibitem[{Pennington et~al.(2014)Pennington, Socher, and
  Manning}]{pennington2014glove}
Jeffrey Pennington, Richard Socher, and Christopher Manning. 2014.
\newblock {GloVe: Global vectors for word representation}.
\newblock In \emph{Proceedings of the 2014 Conference on Empirical Methods in
  Natural Language Processing (EMNLP)}, pages 1532--1543.

\bibitem[{Pil{\'a}n et~al.(2014)Pil{\'a}n, Volodina, and
  Johansson}]{pilan2014rule}
Ildik{\'o} Pil{\'a}n, Elena Volodina, and Richard Johansson. 2014.
\newblock Rule-based and machine learning approaches for second language
  sentence-level readability.
\newblock In \emph{Proceedings of the Ninth Workshop on Innovative Use of NLP
  for Building Educational Applications}, pages 174--184.

\bibitem[{Prince(1996)}]{prince1996second}
Peter Prince. 1996.
\newblock Second language vocabulary learning: The role of context versus
  translations as a function of proficiency.
\newblock \emph{The modern language journal}, 80(4):478--493.

\bibitem[{Ragno(2016)}]{ragno2016use}
Nancy Ragno. 2016.
\newblock \emph{Use the Right Word: Your Quick \& Easy Guide to 158 Words Most
  Often Confused or Misused}.
\newblock Nancy Ragno.

\bibitem[{Ribeiro et~al.(2016)Ribeiro, Singh, and Guestrin}]{ribeiro2016should}
Marco~Tulio Ribeiro, Sameer Singh, and Carlos Guestrin. 2016.
\newblock Why should {I} trust you?: {E}xplaining the predictions of any
  classifier.
\newblock In \emph{Proceedings of the 22nd ACM SIGKDD International Conference
  on Knowledge Discovery and Data Mining}, pages 1135--1144. ACM.

\bibitem[{Room(1988)}]{room1988dictionary}
Adrian Room. 1988.
\newblock \emph{Dictionary of confusing words and meanings}.
\newblock Dorset Press.

\bibitem[{Sinclair(1992)}]{sinclair1992collins}
John Sinclair. 1992.
\newblock \emph{Collins COBUILD English Usage}.
\newblock Collins.

\bibitem[{Tinkham(1993)}]{tinkham1993effect}
Thomas Tinkham. 1993.
\newblock The effect of semantic clustering on the learning of second language
  vocabulary.
\newblock \emph{System}, 21(3):371--380.

\bibitem[{Turton and Heaton(1996)}]{turton1996longman}
Nigel~D Turton and John~Brian Heaton. 1996.
\newblock \emph{Longman Dictionary of Common Errors}.
\newblock Longman.

\bibitem[{Uzawa(1996)}]{uzawa1996second}
Kozue Uzawa. 1996.
\newblock Second language learners' processes of l1 writing, l2 writing, and
  translation from l1 into l2.
\newblock \emph{Journal of second language writing}, 5(3):271--294.

\bibitem[{Walker and Duncan(1967)}]{walker1967estimation}
Strother~H Walker and David~B Duncan. 1967.
\newblock Estimation of the probability of an event as a function of several
  independent variables.
\newblock \emph{Biometrika}, 54(1-2):167--179.

\bibitem[{Waring(1997)}]{waring1997negative}
Robert Waring. 1997.
\newblock The negative effects of learning words in semantic sets: {A}
  replication.
\newblock \emph{System}, 25(2):261--274.

\bibitem[{Webb(2007)}]{webb2007effects}
Stuart Webb. 2007.
\newblock The effects of synonymy on second-language vocabulary learning.
\newblock \emph{Reading in a Foreign Language}, 19(2):120--136.

\bibitem[{Xu and Jordan(1996)}]{xu1996convergence}
Lei Xu and Michael~I Jordan. 1996.
\newblock On convergence properties of the {EM} algorithm for {G}aussian
  mixtures.
\newblock \emph{Neural computation}, 8(1):129--151.

\bibitem[{Yanjuan(2014)}]{yanjuan2014bnc}
HUO Yanjuan. 2014.
\newblock {BNC-Based Design of College English Vocabulary Teaching for Chinese
  College Students}.
\newblock \emph{Studies in Literature and Language}, 8(3):122--125.

\bibitem[{Yeh et~al.(2007)Yeh, Liou, and Li}]{yeh2007online}
Yuli Yeh, Hsien-Chin Liou, and Yi-Hsin Li. 2007.
\newblock Online synonym materials and concordancing for {EFL} college writing.
\newblock \emph{Computer Assisted Language Learning}, 20(2):131--152.

\bibitem[{Zhang and Litman(2018)}]{zhang2018co}
Haoran Zhang and Diane Litman. 2018.
\newblock Co-attention based neural network for source-dependent essay scoring.
\newblock In \emph{Proceedings of the Thirteenth Workshop on Innovative Use of
  NLP for Building Educational Applications}, pages 399--409.

\bibitem[{Zughoul(1991)}]{zughoul1991lexical}
Muhammad~Raji Zughoul. 1991.
\newblock Lexical choice: {T}owards writing problematic word lists.
\newblock \emph{International Review of Applied Linguistics}, 29(1):45--60.

\end{thebibliography}
